\documentclass{article}

\usepackage[english]{babel}

\usepackage[letterpaper,top=2cm,bottom=2cm,left=3cm,right=3cm,marginparwidth=1.75cm]{geometry}

\usepackage{amsmath}
\usepackage{graphicx}
\usepackage[colorlinks=true, allcolors=blue]{hyperref}
\usepackage{booktabs}

\newtheorem{example}{Example}
\newcommand{\nth}[1]{#1^{\text{th}}}

\title{Using Combinatorial Optimization to Design a High quality LLM Solution}
\author{Samuel Ackerman, Eitan Farchi, Rami Katan, Orna Raz}

\begin{document}
\maketitle

\begin{abstract}
We introduce a novel LLM based solution design approach that utilizes combinatorial optimization and sampling.  Specifically, a set of factors that influence the quality of the solution are identified.  They typically include factors that represent prompt types, LLM inputs alternatives, and parameters governing the generation and design alternatives.   Identifying the factors that govern the LLM solution quality enables the infusion of subject matter expert knowledge.  
Next, a set of interactions between the factors are defined and combinatorial optimization is used to create a small subset $P$ that ensures all desired interactions occur in $P$.  Each element $p \in P$ is then developed into an appropriate benchmark.  Applying the alternative solutions on each combination, $p \in P$ and evaluating the results facilitate the design of a high quality LLM solution pipeline. The approach is especially applicable when the design and evaluation of each benchmark in $P$ is time-consuming and involves manual steps and human evaluation.  Given its efficiency the approach can also be used as a baseline to compare and validate an autoML approach that searches over the factors governing the solution.   
\end{abstract}

\section{Introduction}

%TBC Orna - add to the flow the observation that the methodolgy enables the infusion of human knowldge to the quality and design process

Similarly to traditional Machine Learning based solutions, the correctness of Large Language Models (LLM) based solutions which are the focus of this paper is partial by design.  LLMs' forte is in generation.  Thus, differently stated, the LLM solution will sometimes generate incorrect artifacts.  The artifacts generated by the LLM solution may still be useful in spite of the fact that they are not absolutely correct.  A human-in-the-loop approach is typically taken.  A human works with partially correct artifacts and modifies them to achieve a desired goal quicker than with traditional processes.  For example, partially correct code may be generated by the LLM solution and corrected by the programmer through an iterative process of interaction with the LLM solution to develop a desired program.

As LLMs are trained on data, their performance is analyzed as some average of a metric that evaluates the quality of individual generated artifacts.  In this paper we do not focus on metrics and metrics analysis but assume they are given.  Instead, we focus on the design of benchmark's data that is used as input to the LLM solution when analyzing  it. In addition, a key challenge in analysing LLM solutions is the generation of benchmark's data with expected results or labels.  For example, in the case of a code generation solution, each generation data item would be a pair of text input that describes the required code and the desired generated code that the LLM should produce (the expected result).  Again, we do not focus on methods for data preparation but on the design of the requirements of what data needs to be prepared. 

The life cycle of LLM solution development includes -  

\begin{enumerate}
    \item Exploration.  LLMs models are being explored using a variety of prompt alternatives and generation methods to identify the best combination of LLMs model, prompts, and generation parameters for the solution being crafted.  
    \item Analysis.  Benchmarks data and metrics are developed to determine the quality of the LLM solutions and further develop it. 
    \item Control.  The LLM solution performance is monitored as the solution is being used to determine if it works as expected. 
    \item Regression.  The LLM solutions design is revisited and benchmarks are reanalyzed.  This can be triggered either by a new LLM model candidate or a change in usage patterns of the LLM solution.
\end{enumerate}

This work focuses on the analysis and regression stages.  We note that skipping it results in an LLM solution that is based on anecdotal evidence and is most likely to not meet the customers needs. In addition, although the exploration phase is important we advocate moving to the analysis stage as soon as possible.  Thinking about benchmarks as a new type of tests for a new type of systems, namely systems that use LLMs in their constructions, we are basically advocating test first as much as possible of the LLM solution.  The rest of the paper addresses benchmarks design and how it facilitates the design of the LLM solution.  

The main contribution of the paper is in suggesting a methodology and reuse of an existing modeling technology for: 
\begin{enumerate}
    \item Systematically infusing human knowledge in benchmark design,
    \item Providing automated assistance for defining the design space and its coverage, 
    \item Providing technology to efficiently sample the most important variations of the experiments stemming from the design space, and
    \item Providing a method to identify the factors that contribute the most to quality of the results in a specific experimental setup while requiring only minimal data.

\end{enumerate}

\section{Factors that influence the design of the LLM solution}
\label{factors}

In this section we consider the factors that govern a solution that uses LLMs.  Identifying these factors is critical to designing a successful LLM solution.  Through the explicit identification of the governing factors the LLM solution designers are able to infuse knowledge gained through the exploration stage and greatly facilitate the development of the LLM solution.

One or more LLM models, traditional machine learning models and regular software come together to construct an LLM solution.  For example, in a use case of a customer IT problem report, the problem report may be first screened by a traditional ML classifier to determine if the problem pertains to the company's products.  Next, an LLM may be used in combination with a Retrieval Augmented Generation (RAG) pattern to access the company's existing problem repository and construct a potential resolution to the problem.  Finally, another LLM may be used to generate possible scripts that can be run to update the IT configuration and resolve the problem.  In this paper we do not focus on the end-to-end solution design but on the interaction with a specific LLM that is used as part of the end-to-end design. The techniques elaborated here are applicable to the end-to-end use case but the factors to be considered when validating the end-to-end scenario will probably be different.   In addition, the key new component in the LLM solution is the LLM model, hence the focus on the factors that govern the use of the LLM model when incorporated in the end to end system flow.

Once you focus on the interaction with a single LLM to achieve some task such as questions and answers in some given domain or code generation from code there are four categories of factors that govern the LLM solution that need to be considered namely -
\begin{enumerate}
    \item Factors that govern prompt variations
    \item Factors that govern the input to the LLM model
    \item Factors that govern the generation done by the LLM model
    \item Factors that govern the design of the pre processing of the LLM input and post processing of its output.
\end{enumerate}

To facilitate the explanation of these factors we consider as a running example the task of obtaining a summary of some code using an LLM.  In what follows we describe the process of identifying the above four types of factors highlighting the relation of the exploration stage to factor identification.  We also emphasis that this is an ongoing process.  When new LLM models become available, new factors may be identified or old factors dropped through insights gained in the exploration done on the new model.  

First an initial exploration stage against the available models is conducted playing with prompts, inputs and generation parameters.  It is useful to identify at this stage the factors that impact the quality of the solution.  In our running example, the quality of the result depends on the quality of the documentation embedded in the code function headers and body collectively referred to as the inline documentation.  It also depends on whether or not the inline documentation is up to date.  Specifically, in some instances we observe that if the inline documentation is not up to date then, depending on which LLM is used, the summary might be an hallucination.  In addition, if the inline documentation is both up to date and of high quality then it is sometimes just copied as is and the LLM effectively ignores the code in the summary. Both cases are deemed undesired and it becomes clear that an ideal solution should sometimes use the inline documentation and other times not.

To further complicate things it is not clear how to best implement a solution that would take into account the above dependency on inline documentation.  Should we develop a capability to check if the inline documentation is up to date up front?  Alternatively, we could generate code both with and without the inline documentation as input and then choose the best generated summary using some other capability that will be developed possibly by tuning an LLM as a judge?  To even further complicate things it is observed that header and body of inline documentation have different impact on the quality of the summary. Thus four types of inputs should be considered, namely, with the header comments or without, with the body comments or without.

It is also observed that the quality of the summary is impacted by different generation parameters such as the temperature and maximum number of tokens.  In addition, modifications to the prompt may yield better or worse summaries. 

\begin{figure}
\centering
\includegraphics[width=0.8\linewidth]{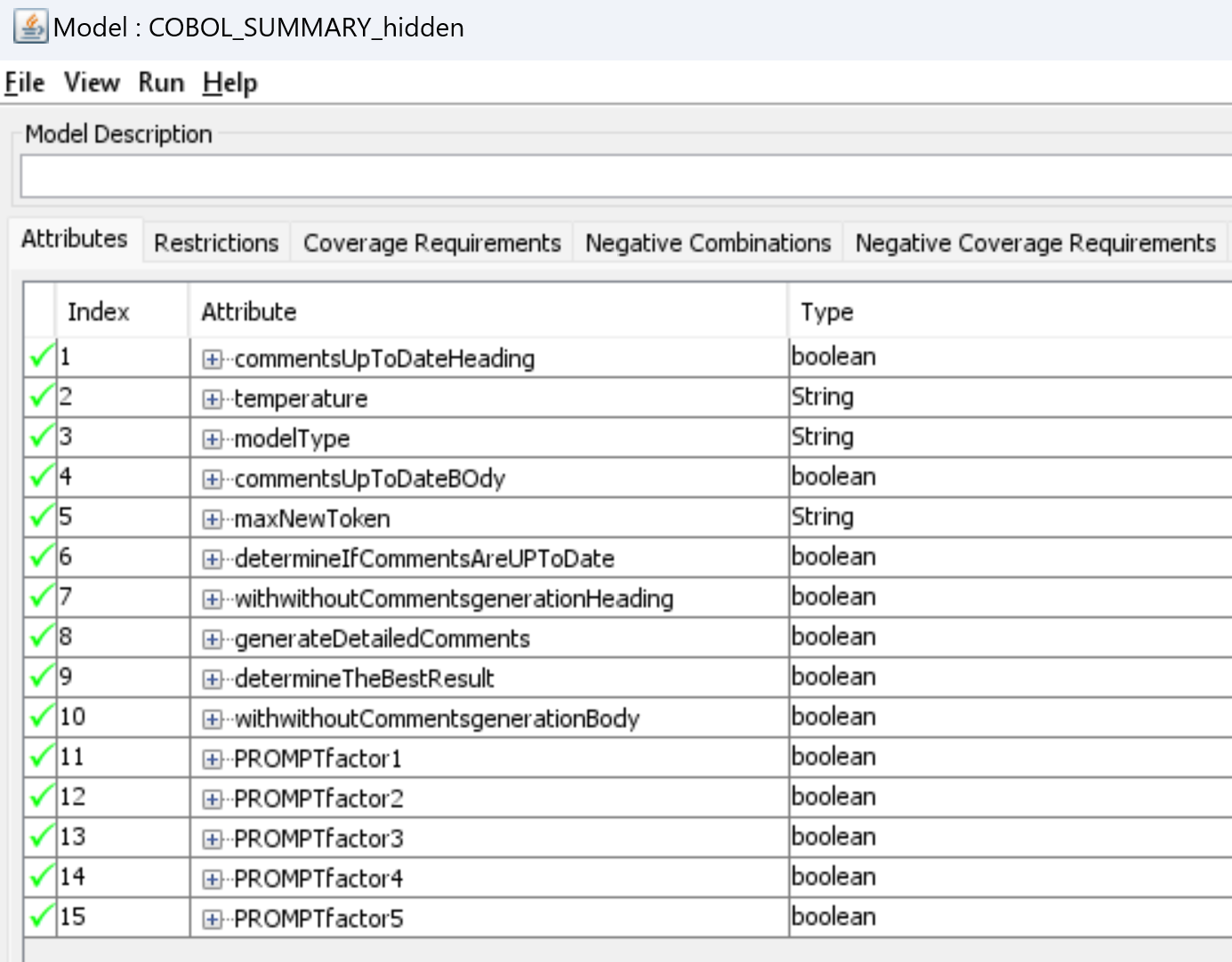}
\caption{\label{fig:parameters}Parameters that impact the quality of the LLM summary solution.}
\end{figure}

The knowledge obtained in the exploration stage on the factors that impact the quality of the LLM solution are captured in a table \ref{fig:parameters}.   The key point to notice is that the design space of the solution includes roughly $200,000$ combinations.   We next discuss how to handle such combinatorial explosion.   
 
\section{The problem of a combinatorial explosion of the design space}
\label{explosition}

%TBC design space of the solution expressed as a Cartesian product of factors
 
Each combination of parameters in \ref{fig:parameters} represents a design alternative.  For example, one combination of parameters may specify the desired format of the code summary in the prompt, use the code documentation, and set the temperature level for the LLM generation at some level while another parameter combinations may generate two candidate summaries, one that uses the code documentation and another that does not and then choose the best summary using another LLM as a judge.  

Given that we have an abundant of code samples, we can readily and reliably estimate the expected average quality of each parameters combinations using human evaluation by using the following procedure.  The procedure below utilizes human data annotation\footnote{The following resource, \href{https://macgence.com/blog/a-brief-guide-about-the-data-annotation/}{link}, may be consulted on details on data annotation techniques and best practices.}.

\begin{enumerate}
    \item Define criteria for a good summary 
    \item Consider a specific parameters value combination from \ref{fig:parameters}
    \item Randomly choose $30$ code samples and apply the required LLM solution alternative determined in the previous step to obtain a summary.  As we are interested in estimating the average and standard deviation of "goodness of the summary" a sample of 20-30 code examples is a good number by the Central limit theorem
    \item Have a human determine the goodness of each code sample and calculate the average and standard deviation of the "goodness of the summary".  In the simple case that we defined only two categories, namely bad summary and good summary, and assuming we have $k$ samples $x_1, \ldots, x_k$ we will end up with $k$ determination of goodness, namely, $g(x_1), \ldots, g(x_k) \in \{1, 0\}$.  $g(x_i)$ is 1 if the human evaluator decided that the summary is good and 0 otherwise 
\end{enumerate}

A back of the envelope calculation indicates that it is not feasible to exhaustively evaluate the table in \ref{fig:parameters} in that manner.  Assuming the length of the code for which we are obtaining the summary is a few pages and assume it will take the human evaluator about $5$ minutes to decide if the summary is good or bad.
Given that we have $200,000$ combinations to analyze, the evaluation will take $200,000 \times 5$ minutes. Too much time!  Augmented with the time it takes to run the solution that match the parameter combinations it becomes evident that it is not reasonable to evaluate each combination in \ref{fig:parameters}.  In the next section we discuss how to overcome the combinatorial explosion we have just highlighted. 

\section{Overcoming the combinatorial explosion}

%Search for relative resources
%https://en.wikipedia.org/wiki/Factor_analysis
%https://en.wikipedia.org/wiki/Exploratory_factor_analysis
%https://en.wikipedia.org/wiki/Factor_analysis

%Add to overleaf of ctd plus LLM
%Sparsity-of-effects principle
%https://en.m.wikipedia.org/wiki/Sparsity-of-effects_principle
% Wu, C. F. Jeff; Hamada, Michael (2000). Experiments: Planning, analysis, and parameter design optimization. New York: Wiley. p. 112. ISBN 0-471-25511-4.

%https://inline.stat.psu.edu/stat503/lesson/5/5.1
%https://en.m.wikipedia.org/wiki/Occam%27s_razor
%Occam’s razor https://en.m.wikipedia.org/wiki/Occam%27s_razor
%Readable clean and understandable design 
%ANOVAhttps://en.m.wikipedia.org/wiki/Analysis_of_variance. 
% Fortunately, experience says that high order interactions are rare
%https://en.m.wikipedia.org/wiki/Two-way_analysis_of_variance
%Two way annova an example
%https://www.wallstreetmojo.com/two-way-anova/

We are facing with the well known situation of a factorial experiment, \href{https://en.wikipedia.org/wiki/Factorial_experiment#:~:text=In%20statistics%2C%20a%20full%20factorial,levels%20across%20all%20such%20factors.}{link},  The sparsity-of-effects principle (see \href{https://en.m.wikipedia.org/wiki/Sparsity-of-effects_principle}{link}), tells us that we should expect the quality of the design of the LLM solution to be dominated by low order one, two or three interactions of parameters. We thus apply the combinatorial optimization described in the next section to obtain a small set that includes all low order interactions of the parameters.  In our running example, the optimized plan will be on the order of ten combinations. Utilizing the procedure described in the previous section for the ten combinations obtained through the combinatorial optimization we choose the parameters combination that has the highest summary quality among the ten parameters combinations chosen in the combinatorial optimization stage.  We also apply Analysis of Variance (ANOVA, see  \href{https://en.wikipedia.org/wiki/Factorial_experiment#:~:text=In%20statistics%2C%20a%20full%20factorial,levels%20across%20all%20such%20factors.}{link}) techniques to identify the small set of parameters that dominate the quality of the code summary.  We next detail the approach.

\section{The combinatorial optimization problem}

We first formalize the optimization problem that we solve.  Consider a set of $n$ attributes, with attribute $i$ having a set $A_i=(a_{i,1},\dots,a_{i,m(i)})$ of possible values (e.g., $A_i=\{a_{i,1}=\textrm{True},\:a_{i,2}=\textrm{False}\}$).  %Given a set of attributes and their values $A = A_1 \times \ldots \times A_n$
The full set of possible combinations of values of the $n$ attributes is $A = A_1 \times \ldots \times A_n$.  Each element of $A$ is an $n$-length element where the $\nth{j}$ element is one of $A_j$.
We would like to create a small set of combinations $P \subseteq A$ that covers a set of attributes values interactions that we are interested in.  For example, $k$-coverage when $k=2$ would mean that for every attribute pair $(i,j),\:i\ne j$, for each $a_i\in A_i$ and $a_j\in A_j$, there exists some $d\in P$, where $d=(d_1, \ldots, d_n)$, so that $d_i=a_i$ and $d_j=a_j$.  That is, each potential such combination $(a_i,a_j)$ occurs at least once (i.e., is covered by the test plan $P$).  As we focus on small order interactions, most of the time, $k$ will be equal to 2 or 3 factors.  In addition, one can require only a subset of the value interactions in $A$ to be covered by $P$. In fact, as part of the methodology we describe below a subset of interactions is chosen by the designer of the experiments based on domain understanding.

Another point is that some combinations may not be possible.  For example, in our running example, if the generation of the LLM is set to a greedy generation then the temperature value may have no impact on the generation.  This is captured using logical constrains that disallow some of the combinations in $A$.  Thus, to be more precise $k-$coverage requires that all combinations of $k$ attribute values that are possible are covered. An example follows.

\begin{example}
Assume that $A_i : Boolean$ is of type Boolean and $n = 4$.  We then have $2^4=16$ combinations.  Assume 
$P \subseteq A$ obtains $2-coverage$, and consider that values $v_1 = 1, v_2 = 1$, then there will be a member of $P$, e.g., $d=(1, 1, 1, 0) \in P$, such that $v_1 = d_1 = 1$ and $v_2 = d_2 = 2$.   This will be true for any combination of two values. 
\end{example}

We next detail the steps taken to design the plan set, $P \subseteq A$

\section{Plan design methodology}

To define the plan set, $P$, the following steps are taken.  We assume that a set $B \subseteq A = A_1 \times \ldots A_n$ is defined of the possible combinations.  This set may be large.  For example, as mentioned in \ref{factors} it could be of the order of $200,000$ combinations or more. We next proceed as follows. 

We define the attributes for which we want to obtain coverage of the values interactions. This is based on human expert knowledge and our methodology and technology explicitly capture this knowledge. For example, in our running example we may want coverage interactions of parameters that define the type of prompt that we are using and parameters that govern the generation done by the LLM.  This will exclude interaction of parameters that govern that input to the LLM model, such as whether or not we are using the code documentation.  We may want to keep them set to some desired value for the current design of the experiment. For example, we may be confident that generating with and without the code and choosing the best generated summary using LLM as a judge gives the best way forward and we are thus do not want to make that as part of the experiment plan but fix these values.  We next obtain a detailed plan $P \subset B$ for our running example.  

%Force the designer to make the choices on the parameters that are not covered  
%Implement as a don’t care constraint? 
%Methodology  
%Define attributes and values 
%For each attribute either  
%Chose to be part of the coverage 
%Add a constrain that specify its value 
%Create the CTD plan 
%For each line in the test plan inspect the attributes that are not coverage and change them if needed 
%Use that to implement the experiments  
%Every combination is now a well thought design alternative  

\section{Running example detailed plan}

\begin{figure}
\centering
\includegraphics[width=1.0\linewidth]{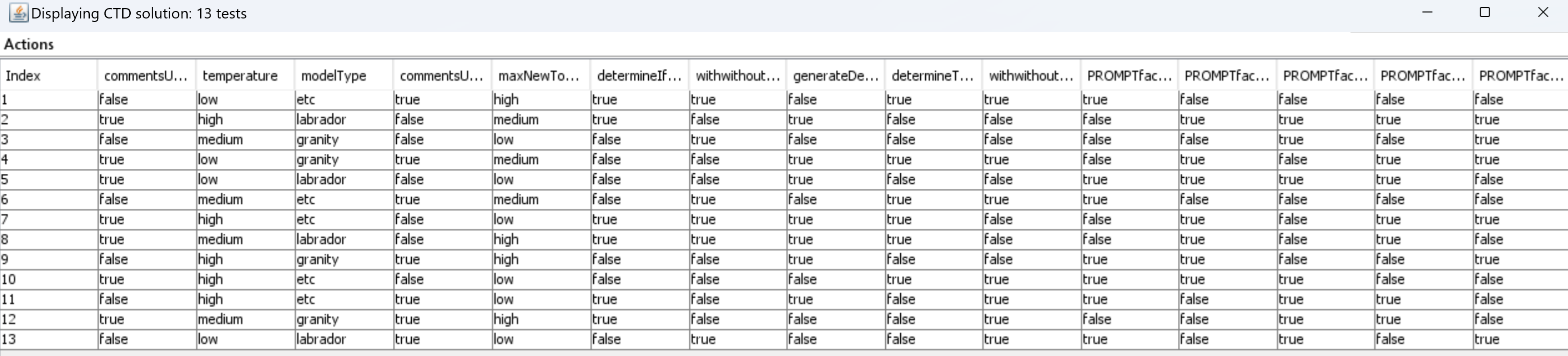}
\caption{\label{fig:CTDPlan}Example of a generated 2-coverage plan of benchmark variations.}
\end{figure}

Utilizing combinatorial optimization we next obtain a subset of combinations (see \ref{fig:CTDPlan}.) that guaranties $2$-coverage, i.e., each possible two values appears in one of the parameters combinations.  For simplicity we are not customizing the $2$-coverage to focus on specific subset of attributes that are of interest.

We thus obtain $12$ combination defining $12$ benchmark variations that need to be implemented.  This benchmarks design enables the analysis of the different dependencies of the parameters and reaching a conclusion of the best solution pipeline including whether to check if the inline documentation is up to date upfront or pick up the best sumerization out of generations alternatives that used the inline documentation or not.  Also it enables the choice of best generation alternatives and best prompting alternatives.  

\section{Analysis}

Once we have scored the generation results on each combination of attribute parameter values (elements in the plan set $P=(p_1,\dots,p_m)$), we attempt to determine the best possible combinations of attributes to be used in the solution and identify which attributes are dominating its quality. To that end we conduct a statistical analysis that has two main objectives:

\begin{enumerate}
    \item Evaluate whether the average scores between various combinations are statistically significantly different.
    \item Evaluate the relative influence of the attributes and their levels on the score.
\end{enumerate}

Realizing the first objective above will help us determine the best combination in $P$ (or in \ref{fig:CTDPlan} for our running example).  The second objective will help us gain insights on the interactions of parameters values that are influencing the quality of the solution.

\subsection{Finding the highest-scoring combination of attribute values
\label{ssec:best_combination}
}

The first task involves assessing the statistical significance of the relative average score between various combinations.  Here, we consider the scores resulting from each unique attribute combination in $P$ as comprising a sample $S_i$; so say $|P|=12$ unique combinations give us $|P|$ samples $S_1,\dots,S_{|P|}$.  Naturally, the best attribute combination among $P$, judging by the evidence of the collected data, is the one for which the sample average $\bar{S}_i$ is highest.  However, we would like to determine whether the differences in average scores between samples are statistically significant (i.e., more than would be expected under random chance).  For instance, letting $a$ and $b$ denote the highest and second-highest values of the sample averages, we may want to determine if $\bar{S}_a$ is significantly higher than $\bar{S}_b$ before recommending the first combination as the best.  The procedure below recommends how to properly collectively test \emph{every pair} $\bar{S}_i$ vs $\bar{S}_j$ for every $(i,j)$ where $i=1,\dots,|P|-1$ and $j=i+1,\dots,|P|$.  In total, there are $\binom{|P|}{2}$ such unique pairs.

Statistically, each observed score in sample $S_i,\:i=1,\dots,|P|$ is considered to theoretically follow a Bernoulli distribution $B(\theta_i)$ with a particular unknown proportion value $\theta_i\in[0,1]$. The sample average $\bar{S}_i$ should be $\approx \theta_i$. Since each code instance is independently and uniformly drawn from the same dataset, the average expected score ($\theta_i$) depends only on the attribute value combination that yielded the sample values $S_i$, with some configurations more likely to score higher than others. 

For every pair $(i,j)$, we can conduct a \href{https://www.statsmodels.org/dev/generated/statsmodels.stats.proportion.proportions_ztest.html}{two-sample proportions T-Test}, with null hypothesis $H_{0;\:i,j}\colon\:\theta_i-\theta_j=0$ vs the alternative $H_{A;\:i,j}\colon\:\theta_i-\theta_j\ne 0$.  This determines whether, on the basis of the observed sample means for the two attribute combinations $i,j$, whether their scores appear significantly different. Each such test yields a p-value $p_{i,j}\in[0,1]$.  This set of $\binom{|P|}{2}$ total p-values is adjusted by a \href{https://www.statsmodels.org/dev/generated/statsmodels.stats.multitest.multipletests.html}{multiple testing procedure} such as Holm-Sidak.  We can then look at any adjusted p-value $\tilde{p}_{i,j}$ for a comparison of two configurations' score samples.  If this $\tilde{p}_{i,j}$ is low (e.g. lower than $\alpha=0.05$), then the two configurations have statistically significantly different average scores $\theta_i,\theta_j$.

We note that this approach can determine statistical significance of the scores only for the particular test plan combinations considered.  It does not attempt to conduct an optimization by inferring on unseen combinations.  In some cases, it may be sufficient to identify the best combination in the $P$ considered, if it is good enough for the user's purpose.

\subsection{Modeling the effect of attribute values on the score
\label{ssec:attribute_score_effect}
}

Unlike the first task (Section~\ref{ssec:best_combination}), in this task we propose to build a predictive model on the data, rather than only directly using the score samples $S_1,\dots,S_{|P|}$. This predictive model will let us statistically evaluate the effect of any attribute and its values on the score, and optionally find the optimal attribute value combination (possibly not in $P$).

We propose to  formulate a logistic regression model where the attributes are the explanatory variables and the (binary) score of each instance of the combination is the target variable.  This model may also include interactions (say, of up to order 2) between the attributes.  Analysis of the regression results compromises several steps:

\begin{itemize}
    \item Within each attribute, assess the average influence of each level on the target (particularly for numeric variables and categorical ones with more than two levels---currently only temperature, modelType, and maxNewToken).  I.e., does this value positively or negatively affect the score, and is it a significant effect.  This is done by examining the p-value of the regression coefficient and the odds-ratio values.
    \item Examining the Wald table (an output of fitting a logistic model) to assess the \emph{overall} statistical contribution of each attribute (or interaction of attributes) to the model in terms of explaining the target.  The lower the attribute p-value, the more significant it is; thus we can determine which attributes do or don't affect the score.
\end{itemize}

See \cite{rabinovich-etal-2023-predicting} for examples of interpretation of the regression outputs.  For this regression model to successfully compute the effects of each attribute, the plan needs to have enough \emph{different} combinations of parameter values (not total evaluation runs per combination), relative to the number of attributes (and their interactions) considered.  In particular, $P$ will need to be larger if we want to model attribute interactions in addition to single-attribute effects.  For instance, the two-attribute combination (commentsUpToDateHeading="True") $\:\&\:$ (temperature="low") needs to appear as a subset of several different larger attribute combinations, rather than in only one, in order to statistically evaluate the effect of this value combination in isolation from the values of other attributes.  
\subsection{Analysis simulation
\label{ssec:analysis_simulation}
}

Here we show a brief simulation with synthetic data.
First, a set of 265 unique test plans $P=(p_1,\dots,p_{265})$ are created by 20 independent generations of a test plan using the combinatorial optimization, each of which guarantees coverage of each attribute value pair.  As mentioned in Section~\ref{ssec:attribute_score_effect}, this repetition is necessary to obtain enough examples of each fixed 2-attribute combination with other attribute values so that the effect of each value on the score can be analyzed by logistic regression.  For each plan $p_i$, scores are generated by a hierarchical simulation of $\alpha_i\sim \textrm{Gamma}(5,1)$ and $\beta_i\sim \textrm{Gamma}(2,1)$, with $\theta_i\sim \textrm{Beta}(\alpha_i, \beta_i)$. Each test plan $p_i$ then has a sample $S_i$ of 30 scores generated independently from the distribution $\textrm{Ber}(\theta_i)$.  This is done to give each plan a random but different mean accuracy $\theta_i$.  Again, on average $\bar{S}_i=\theta_i$.  Thus, our dataset $D$ consists of $265\times30=7950$ instances.  The number $n$ of samples per combination could be reduced to save effort, but it could reduce the statistical significance of results.  

Here, for simplicity, we only show the regression results of the score on each of the attributes, without interactions between them.  In all tables, significance is coded by the `symbol' column, which codes\footnote{This notational convention is used in \texttt{R} statistical software.} the statistic's p-value: *** ($<0.001$), ** ($<0.01$), * ($<0.05$), . ($<0.1$), or blank ($\geq 0.1$).  Table~\ref{tab:sim_regression_coeffs} shows the value of each categorical regression variable, and its statistical significance.  Thus, for instance, a plan for which maxNewToken=``medium" yields a relatively significantly lower score (by 0.115 on average) than if maxNewToken=``low" (the baseline level); this has a p-value of 0.058.  In contrast, the effect of maxNewToken=``high" is essentially the same as ``low", since the p-value is 0.959 (not significant at all). The results may differ if we analyze interactions.

\begin{table}[h]
\begin{tabular}{lrrrrl}
\toprule
 & coefficient & [0.025 & 0.975] & p-value & symbol \\
\midrule
Intercept & 0.543 & 0.330 & 0.756 & 0.000 & *** \\
C(commentsUpToDateHeading)[T.True] & 0.005 & -0.094 & 0.104 & 0.918 &  \\
C(temperature)[T.medium] & 0.286 & 0.160 & 0.413 & 0.000 & *** \\
C(temperature)[T.high] & -0.084 & -0.202 & 0.034 & 0.165 &  \\
C(modelType)[T.granity] & 0.283 & 0.161 & 0.404 & 0.000 & *** \\
C(modelType)[T.labrador] & 0.371 & 0.254 & 0.489 & 0.000 & *** \\
C(commentsUpToDateBOdy)[T.True] & 0.105 & 0.005 & 0.205 & 0.040 & * \\
C(maxNewToken)[T.medium] & -0.115 & -0.233 & 0.004 & 0.058 & . \\
C(maxNewToken)[T.high] & -0.003 & -0.127 & 0.120 & 0.959 &  \\
C(determineIfCommentsAreUPToDate)[T.True] & -0.183 & -0.284 & -0.082 & 0.000 & *** \\
C(withwithoutCommentsgenerationHeading)[T.True] & 0.127 & 0.026 & 0.228 & 0.014 & * \\
C(generateDetailedComments)[T.True] & 0.132 & 0.032 & 0.232 & 0.010 & ** \\
C(determineTheBestResult)[T.True] & 0.097 & -0.004 & 0.198 & 0.059 & . \\
C(withwithoutCommentsgenerationBody)[T.True] & -0.154 & -0.256 & -0.053 & 0.003 & ** \\
C(PROMPTfactor1)[T.True] & 0.015 & -0.085 & 0.116 & 0.766 &  \\
C(PROMPTfactor2)[T.True] & 0.126 & 0.024 & 0.227 & 0.015 & * \\
C(PROMPTfactor3)[T.True] & 0.073 & -0.028 & 0.174 & 0.157 &  \\
C(PROMPTfactor4)[T.True] & 0.155 & 0.052 & 0.259 & 0.003 & ** \\
C(PROMPTfactor5)[T.True] & -0.089 & -0.190 & 0.011 & 0.082 & . \\
\bottomrule
\end{tabular}
\caption{Coefficient testing of logistic regression results on simulated data, no interaction between attributes.
\label{tab:sim_regression_coeffs}}
\end{table}

Table~\ref{tab:sim_regression_wald} shows the Wald test statistics, which measures the contribution of each of the attributes toward explaining the variation in the target variable (the score).  Unlike Table~\ref{tab:sim_regression_coeffs}, this table does not separate the explanatory power of each attribute level (value of $A_i$), nor does it show the direction of the effect.  As before, an attribute with a low p-value has a strong effect on the score.  In this simulation, the most important attributes are
modelType, temperature, determineIfCommentsAreUPToDate.  Table~\ref{tab:sim_regression_coeffs}, however, can tell us whether the influence is due to only one of the attribute levels or whether all levels have a differential influence.

\begin{table}
\begin{tabular}{lrrrl}
\toprule
 & chi2 & p-value & df constraint & symbol \\
\midrule
Intercept & 24.876 & 0.000 & 1 & *** \\
C(commentsUpToDateHeading) & 0.011 & 0.918 & 1 &  \\
C(temperature) & 35.675 & 0.000 & 2 & *** \\
C(modelType) & 42.081 & 0.000 & 2 & *** \\
C(commentsUpToDateBOdy) & 4.218 & 0.040 & 1 & * \\
C(maxNewToken) & 4.616 & 0.099 & 2 & . \\
C(determineIfCommentsAreUPToDate) & 12.725 & 0.000 & 1 & *** \\
C(withwithoutCommentsgenerationHeading) & 6.022 & 0.014 & 1 & * \\
C(generateDetailedComments) & 6.658 & 0.010 & 1 & ** \\
C(determineTheBestResult) & 3.573 & 0.059 & 1 & . \\
C(withwithoutCommentsgenerationBody) & 8.896 & 0.003 & 1 & ** \\
C(PROMPTfactor1) & 0.089 & 0.766 & 1 &  \\
C(PROMPTfactor2) & 5.891 & 0.015 & 1 & * \\
C(PROMPTfactor3) & 2.003 & 0.157 & 1 &  \\
C(PROMPTfactor4) & 8.652 & 0.003 & 1 & ** \\
C(PROMPTfactor5) & 3.025 & 0.082 & 1 & . \\
\bottomrule
\end{tabular}
\caption{Wald table of logistic regression results on simulated data, no interaction between attributes.
\label{tab:sim_regression_wald}}\end{table}

Say we now consider only the combinations $P$ from the first generation of the 2-coverage plan (Figure~\ref{fig:CTDPlan}), which consists of 13 sets of parameters values.  The results show that plan $p_{2}\in P$ has the best score ($\bar{S}_{2}=1.0$), with $p_7$ being second ($\bar{S}_7=0.9667$).  The raw p-value of the test between these is 0.313, which becomes 1.0 after multiplicity adjustment (since we did not know a priori we would be interested in this particular pair).  This high p-value means that the difference between these two attribute configurations is not statistically significant, given the sample sizes.

\section{Related work}

AutoML \cite{AutoML}, and more recently 
AutoRAG \cite{AutoRAG}, that applies to generative AI use cases, helps identify the best combination of parameters including models and their hyper parameters, prompts variations, and input alternatives.  Typically, these approaches assume that you can quickly and automatically decide on the performance of each specific parameters combination.  In this work this assumption does not apply as evaluation of a single pipeline is long and partially manual due to inference time as well as lack of appropriate metrics. 

\section{Conclusion}

We have outlined how factors that are found to impact the LLM solutions are captured using attributes and values and how combinatorial optimization and statistical analysis is applied to choose a high quality combination of parameters values and systematically create a high quality solution.  From time to time new LLM models become available.  At that point the analysis process can be efficiently repeated to re-validate or change the choice of parameters values that govern the LLM solution quality. 

\bibliographystyle{alpha}
\bibliography{sample}

\end{document}